\pdfoutput=1

\documentclass[11pt]{article}

\usepackage{ACL2023}

\usepackage{times}
\usepackage{latexsym}

\usepackage[T1]{fontenc}

\usepackage[utf8]{inputenc}

\usepackage{microtype}

\usepackage{inconsolata}
\usepackage{ulem}

\title{\aramus{}: Pushing the Limits of Data and Model Scale for Arabic Natural Language Processing}

\author{Asaad Alghamdi${^1,}\thanks{\hspace{2mm}Equal contribution}$\hspace{3mm}Xinyu Duan${^2,}$\footnotemark[1]\hspace{3mm}Wei Jiang$^2$ \hspace{2mm}Zhenhai Wang$^2$\hspace{2mm}Yimeng Wu$^3$\\
\textbf{Qingrong Xia$^2$\hspace{2mm}Zhefeng Wang$^2$\hspace{2mm}Yi Zheng$^2$\hspace{2mm}Mehdi Rezagholizadeh$^3$\hspace{2mm}Baoxing Huai$^2$}\\
\textbf{Peilun Cheng$^1$\hspace{2mm}Abbas Ghaddar$^3$}\\
$^1$ AI Cognitive Team, Tonomus\\
$^2$ Huawei Cloud Computing Technologies Co., Ltd. \\
$^3$ Huawei Technologies Co., Ltd.\\
\texttt{\{asaad.alghamdi,eddie.chengpeilun\}@neom.com}\\
\texttt{\{duanxinyu,jiangwei160,wangzhenhai1,yimeng.wu,xiaqingrong,wangzhefeng,}\\
\texttt{zhengyi29,mehdi.rezagholizadeh,huaibaoxing,abbas.ghaddar\}@huawei.com}\\
}

\usepackage{amssymb}
\usepackage{pifont}

\newcommand{\bert}{BERT}

\newcommand{\cmlbert}{CAMeLBERT} 
\newcommand{\arbert}{ARBERT}
\newcommand{\marbert}{MARBERT}
\newcommand{\jaber}{JABER}

\newcommand{\saber}{SABER}

\newcommand{\ats}{AT5S}
\newcommand{\atb}{AT5B}
\newcommand{\artb}{AraT5-base}

\newcommand{\aramus}{AraMUS}

\newcommand{\mightmention}[1]{}
\newcommand{\problem}[1]{\textcolor{red}{$\star$}}
\newcommand{\answer}[1]{\textcolor{blue}{$\#$}}
\newcommand{\todoreview}[1]{\textcolor{green}{$@$}}

\usepackage{subcaption}
\usepackage{rotating,csquotes}
\usepackage{xcolor}
\usepackage{multirow}
\usepackage{framed}
\usepackage{booktabs}
\usepackage{amsmath}
\usepackage{upgreek}
\usepackage{ulem}
\usepackage{amsfonts}

\usepackage[most]{tcolorbox}

\newtcbox{\mybox}[1][]{enhanced, colframe=blue, colback=blue!15, 
	frame style={opacity=0.25}, interior style={opacity=0.25}, 
	nobeforeafter, tcbox raise base, shrink tight, extrude by=1mm, #1}

\begin{document}
\maketitle

\begin{abstract}

Developing monolingual large  Pre-trained Language Models (PLMs) is shown to be very successful in handling different tasks in Natural Language Processing (NLP). In this work, we present \aramus{}, the  largest Arabic PLM with 11B parameters trained on 529GB of high-quality Arabic textual data. \aramus{} achieves state-of-the-art performances on a diverse set of Arabic classification and generative tasks. Moreover, \aramus{} shows impressive few-shot learning abilities compared with the best existing Arabic PLMs. 

\end{abstract}

\section{Introduction}
\vspace*{-2mm}

Scaling-up Pre-trained Language Models (PLMs) has led to astonishing performance gains on a vast variety of Natural Language Processing (NLP) tasks~\cite{glamGoogle,zoph2022designing,smith2022using}. It has also opened new perspectives for studying the opportunities and limitations of large PLMs~\cite{raffel2019exploring,dale2021gpt,bommasani2021opportunities}, as well as their social and ethical impacts ~\cite{bender2021dangers,weidinger2021ethical,tamkin2021understanding,rae2021language,susnjak2022chatgpt}.

Although for some languages such as English and Chinese, several PLMs with even more than hundred billions of parameters have been developed~\cite{rae2021scaling,chowdhery2022palm,zeng2021pangu,sun2021ernie}, little or no progress has been made on this direction for many other languages including Arabic.\footnote{Arabic is among top 10 most popular languages in the world with 420M native speakers, and more than 25 popular dialects~\cite{guellil2021arabic}.} 
While there have recently been few attempts to develop multi-billion parameters Arabic PLMs \cite{nagoudi2022jasmine,antoun2021aragpt2,lakim2022holistic}, still, their performances and abilities have not been well investigated. The largest well-studied Arabic PLM has no more than 370M parameters~\cite{nagoudi2022_arat5,ghaddar2022revisiting}.

In this work, we introduce \aramus{}, an 11B parameter encoder-decoder T5~\cite{raffel2019exploring} model, which is pre-trained on 529GB of high-quality Arabic text (filtered out of 8.8TB). To the best of our knowledge, \aramus{} is the largest Arabic PLM in terms of pre-training data and model size. Furthermore, it is the first time a multi-billion parameter Arabic PLM is \textit{systematically} evaluated, against the existing state-of-the-art models, on a diversified set of  discriminative and generative task models. More precisely, \aramus{} achieves new state-of-the-art performances of 79.8\% on the  ALUE~\cite{seelawi2021alue} benchmark, which is a collection of 8 discriminative tasks. In addition, it significantly outperforms the best available encoder-decoder models on multiple generative tasks. Finally, \aramus{} shows remarkable abilities to maintain its performance under few-shot settings.

\section{Related Work}
\label{sec:Related Work}
\vspace*{-2mm}

Recently, there has been a growing body of the literature on very large-scale English PLMs by thoroughly studying different aspects of their scaling. These efforts can be summarized into scaling their  pre-training data~\cite{hoffmann2022training} and model size~\cite{dale2021gpt,rae2021scaling,smith2022using}, designing efficient architectures~\cite{zoph2022designing,chowdhery2022palm} and pre-training objectives~\cite{bajaj2022metro,tay2022unifying}, democratizing their access~\cite{zhang2022opt}, and making them useful in real-world applications~\cite{ouyang2022training,qu2023survey}. Besides English, there have been multiple attempts to develop multilingual~\cite{scao2022bloom}, as well as non-Anglocentric ~\cite{zeng2021pangu,sun2021ernie,shin2022effect}  multi-billion PLMs.

Unfortunately, the development of Arabic PLMs does not follow the same pace as that of English. The earliest released Arabic PLMs~\cite{antoun2020arabert,safaya2020kuisail} were based on the \bert{\textit{-base}} (as well as \textit{-large}) architecture~\cite{devlin2018bert} and pre-trained on less than 100GB of unfiltered data. Successive works tried to improve Arabic \bert{-base} models performance by scaling up the pre-training data up to 197GB and 167GB of unfiltered Arabic text for \marbert{}~\cite{abdul2021arbert} and \cmlbert{}~\cite{inoue2021interplay} respectively. In addition, other works focused on developing Arabic PLMs to support other architectures like AraElectra~\cite{antoun-etal-2021-araelectra}, AraGPT~\cite{antoun2021aragpt2},
AraT5~\cite{nagoudi2022_arat5}, and AraBART~\cite{eddine2022arabart} which are equivalent to English ELECTRA~\cite{clark2020electra},
GPT~\cite{radford2018improving_gpt},
T5~\cite{raffel2019exploring}, and BART~\cite{lewis2019bart} respectively.

Recently, \citet{ghaddar2022revisiting} developed state-of-the-art Arabic \bert{} (\jaber{} and \saber{}) and T5 models (\ats{} and \atb{}) by improving the pre-training data quantitatively and qualitatively. More precisely, they pre-trained Arabic \bert{-base/large} and T5-small/base models on  115GB of high-quality Arabic text data (filtered out of 514GB). AraGPT-Mega~\cite{antoun2021aragpt2}, Jasmine~\cite{nagoudi2022jasmine},  NOOR~\cite{lakim2022holistic} are the only existing multi-billion Arabic PLMs. These are decoder-only GPT models with 1.5B, 6.7B, and 10B parameters respectively. However, these aforementioned works suffer from the absent (e.g. in AraGPT, NOOR) or limited (e.g. Jasmine) comprehensive evaluation on NLP end-tasks. Moreover, some of these models (such as NOOR and Jasmine) are not publicly available for custom evaluations.\footnote{We refer the reader to Appendix~\ref{sec:Baseline} for detailed positioning of \aramus{} against each of these three models.} Evaluation is a key factor for understanding the strengths and limitations of these models, without which the progress of the Arabic NLP field is hindered.

\section{\aramus{}}
\label{sec:Experiments}
\vspace*{-2mm}
\subsection{Pre-training Data}
\vspace*{-1mm}

We mainly leverage all (up to July 2022) of the 90
Common Crawl~\footnote{\url{https://commoncrawl.org}} monthly web scrapes in order to collect massive amount of Arabic textual data. This is significantly larger compared to \jaber{}~\cite{ghaddar2022revisiting}, NOOR~\cite{lakim2022holistic}, and Jasmine~\cite{nagoudi2022jasmine}, which use 10, 21, and 71 monthly CC shards, respectively. Then, we apply aggressive noise filtering and deduplication, which give rise to 529GB of high-quality Arabic text data. \citet{nagoudi2022jasmine} introduced the closest comparable pre-training corpus size to us with 413GB (22\% smaller than ours) of Arabic text data. Our data mainly differs in using 2.5 times more CC data, while they used 3.8 times more dialect data than ours. We refer the reader to Appendix~\ref{sec:Pretraining Data Collection} for technical details regarding the pre-training data collection.

\begin{table*}[!htp]

\centering
\resizebox{\textwidth}{!}{
\begin{tabular}{l|c|cccccccc|c}
\toprule
\textbf{Model} & \textbf{\#Params} & \textbf{MQ2Q}& \textbf{MDD}& \textbf{SVREG}& \textbf{SEC}& \textbf{FID}& \textbf{OOLD}& \textbf{XNLI}& \textbf{OHSD} & \textbf{Avg.}\\

\midrule
\multicolumn{10}{c}{\textit{\bert{-models}}}\\
\midrule
 
\arbert & 163M & 74.7$\pm$0.1 & 62.5$\pm$0.2 & 83.5$\pm$0.6 & 43.9$\pm$0.6 & 85.3$\pm$0.3 & 90.5$\pm$0.5 & 70.8$\pm$0.5 & 81.9$\pm$2.0 & 74.1$\pm$0.6 \\
\marbert& 163M  & 69.1$\pm$0.9 & 63.2$\pm$0.3 & 88.0$\pm$0.4 & 47.6$\pm$0.9 & 84.7$\pm$0.4 &  91.8$\pm$0.3 & 63.3$\pm$0.7 & 83.8$\pm$1.4 & 73.9$\pm$0.7\\
\jaber & 135M  & 75.1$\pm$0.3 & 65.7$\pm$0.3 & 87.4$\pm$0.7 & 46.8$\pm$0.8 & 84.8$\pm$0.3 & 92.2$\pm$0.5 & 72.4$\pm$0.7 & 85.0$\pm$1.6 & 76.2$\pm$0.7 \\
\saber & 369M & 77.7$\pm$0.4 & 67.4$\pm$0.2 & 89.3$\pm$0.3 & 49.0$\pm$0.5 &  86.1$\pm$0.3 & 93.4$\pm$0.4 &  75.9$\pm$0.3 & \textbf{88.9$\pm$0.3} & 78.5$\pm$0.3 \\

\midrule
\multicolumn{10}{c}{\textit{T5-models}}\\
\midrule

\atb  & 296M     & 73.7$\pm$0.1 & 64.7$\pm$0.2 & 78.1$\pm$2.4     & 43.8$\pm$0.7     &  83.1$\pm$0.5 & 90.0$\pm$0.4     & 72.2$\pm$0.4     & 81.2$\pm$2.1 & 73.3$\pm$0.9 \\

\artb & 289M  & 70.5$\pm$2.1 & 63.6$\pm$0.2     & 80.8$\pm$1.3 & 44.0$\pm$0.6 & 82.3$\pm$0.4     & 90.5$\pm$0.4 & 72.5$\pm$1.5 & 78.3$\pm$1.4     & 73.0$\pm$1.0  \\
\aramus{} & 11B  & \textbf{80.7$\pm$0.1} & \textbf{68.0$\pm$0.2} & \textbf{89.8$\pm$0.3} & \textbf{49.6$\pm$0.7} & \textbf{86.6$\pm$0.4} & \textbf{93.8$\pm$0.4} & \textbf{82.9$\pm$0.2} & 
88.2$\pm$1.0 &
\textbf{79.9$\pm$0.2}   \\    
\bottomrule

\end{tabular}}
\caption{\textsc{Dev} set performances and standard deviations over 5 runs on the ALUE benchmark.}
\label{tab:dev_alue}
\end{table*}

\begin{table*}[!htp]
\small
\centering
\resizebox{\textwidth}{!}{
\begin{tabular}{l|c|cccccccc|c||c}
\toprule
\textbf{Model} & \textbf{\#Params} & \textbf{MQ2Q}& \textbf{MDD}& \textbf{SVREG} & \textbf{SEC}& \textbf{FID}& \textbf{OOLD} & \textbf{XNLI}& \textbf{OHSD} & \textbf{Avg.}& \textbf{DIAG}\\
\midrule

\jaber & 135M  & 93.1 & 64.1 & 70.9 & 31.7 & 85.3 & 91.4 & 73.4 & 79.6 & 73.7  & 24.4\\ 
ALM-1.0 & 350M  & 94.5 & 65.1 & 70.1 & 35.3 & 86.0 & 91.7 & 77.7 & 85.7 & 75.8 & 30.2 \\
\saber & 369M  & 93.3 & 66.5 & 79.2 & 38.8 & 86.5 & 93.4 & 76.3 & 84.1 & 77.3  & 26.2\\ 
\midrule
\artb & 282M   & 91.3 & 63.8 & 65.9 & 30.5 & 82.3 & 88.8 & 68.2 & 77.9 & 71.1 & 15.4 \\ 
\aramus{} & 11B   &\bf 95.2 & \bf 67.5 &\bf 80.4 & \bf 41.6 & \bf 87.2 & \bf 95.5 & \bf 83.2 & \bf 87.4 & \bf 79.8 & \bf 42.0 \\ 
\bottomrule

\end{tabular}
}
\caption{Results of top-ranked models on the ALUE leaderboard.}
\label{tab:test_alue}
\end{table*}

\subsection{Model and Implementation}
\vspace*{-1mm}

\aramus{} follows the same encoder-decoder architecture and configuration as T5-xxl~\cite{raffel2019exploring} model with  
64k vocabulary size. We choose encoder-decoder T5 architecture because it was found to deliver a good balance between the performance of the discriminative and generative tasks~\cite{raffel2019exploring,tay2022unifying}, compared to encoder-only \bert{} (discriminative tasks focused) and decoder-only GPT~\cite{radford2019language} 
(generative tasks focused). \aramus{} has 11B parameters in total, which makes it the largest existing Arabic T5 model. It was pre-trained using 128 NVIDIA A100 GPUs for 2 months. Technical details regarding implementation and hyper-parameters used for pre-training are listed in Appendix~\ref{sec:Pretraining Implementation Details}.

\subsection{Evaluation Protocol}
\vspace*{-1mm}
We assess \aramus{} by performing extensive fine-tuning experiments on a diverse set of NLP tasks. On one side, we experiment on 8 tasks from the well-established ALUE benchmark~\cite{seelawi2021alue}, which includes one regression (SVREG), one multi-label classification (SEC), 4 single-sentence (MDD, FID, OOLD, and OHSD) and 2 sentence-pair (MQ2Q and XNLI) classification tasks. On the generative tasks side, we evaluate on Question Answering (QA), Question Generation (QG), and Text Summarization (TS).

We compare \aramus{} with state-of-the-art Arabic PLMs in the literature, including \arbert{}, \marbert{}, \jaber{} (\bert{-base}), \saber{}, ALM-1.0 (\bert{-large}), \atb{} and \artb{}  (T5-base). The experimental protocol is designed to ensure the diversity of the tasks, and the public availability of models. Most importantly, we make sure that datasets are of high quality, open-sourced, and supported by a well-established evaluation protocol. Our goal is to have a fair comparison between models, as well as the credibility and reproducibility of the results. A detailed description of fine-tuning datasets, evaluation metrics, baselines, and  implementation details are available in Appendix~\ref{sec:Finetuning}.

\begin{figure*}[!htb]
    \centering
    \includegraphics[width=16cm]{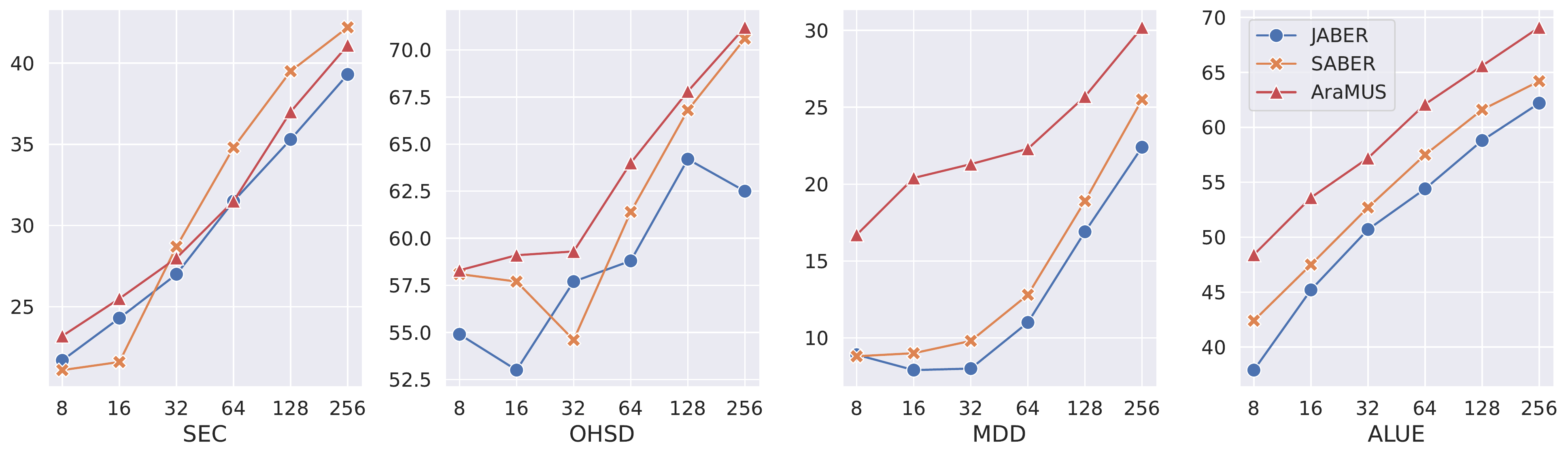}
    \caption{Models performance on the dev set of 3 ALUE tasks and the ALUE average score in the few-shot setting.}
    \label{fig:fewshot_alue}
\end{figure*}

\subsection{Results}
\label{sec:Results}
\vspace*{-1mm}

Table~\ref{tab:dev_alue} shows the dev set results of the eight ALUE tasks with their average scores and standard deviations of 5 runs. The baseline results are directly brought from \cite{ghaddar2022revisiting} and they are directly comparable with \aramus{} since we follow the same evaluation protocol. Table~\ref{tab:test_alue} shows the test set performances of the state-of-the-art models on the ALUE leaderboard. 

As we expect, \aramus{}  outperforms all other baseline models on both dev and test sets and achieves a new state-of-the-art performances on ALUE. 
While our average ALUE result is 1.4\% better than the best baseline, \saber{}, the latter outperforms \aramus{} on the OHSD dataset. 
On the other hand, \aramus{} significantly outperforms \saber{} by 2.5\% on average and 3.3\% on OHSD when comparing results on the leaderboard test. Interestingly, this is roughly a similar performance gap (2.1\%) on the English GLUE~\cite{wang2018glue} between the English T5-xxl~\cite{raffel2019exploring} (11B parameters) and the well-trained  English Roberta-large~\cite{liu2019roberta} model. 

Moreover, we observe a huge gap of 13.8\% between \aramus{} and \saber{} on the ALUE diagnostic set. DIAG was specifically designed to evaluate models' abilities to capture complex linguistic phenomena in Arabic~\cite{seelawi2021alue}. These observations clearly indicate that scaling the model with more data and parameters greatly improves the robustness and generalization abilities of Arabic PLMs. It is worth mentioning that our results are in contrast with previous observations reported in~\cite{nagoudi2022_arat5,ghaddar2022revisiting} that encoder-decoder T5 architecture Arabic models (e.g. \artb{} and \atb{}) significantly underperform \bert{} models on discriminative tasks. Our results suggest that, for Arabic, encoder-decoder models require more data and parameters to catch up with encoder-only models on discriminative tasks.

\begin{table}[!ht]
\centering
\resizebox{\columnwidth}{!}{
    \begin{tabular}{l|cc|cc}
    \toprule
    
    & \multicolumn{2}{c|}{\textit{\bf Dev}} & \multicolumn{2}{c}{\textit{\bf Test}} \\
    \bf Model & \bf EM & \bf F1 & \bf EM & \bf F1\\
    \midrule
    \artb & 40.2$\pm$0.4 & 61.4$\pm$0.8 & 31.2 & 65.7 \\
    \atb & 40.8$\pm$0.7 &  61.6$\pm$1.1 & 31.6 & 67.2 \\
    \midrule
    \aramus & \bf49.8$\pm$1.1 & \bf 69.1$\pm$0.9 & \bf 35.3 & \bf 72.3\\
    \bottomrule
    \end{tabular}
    }
    \caption{F1-score and Exact Match (EM) scores of T5-style models on the Question Answering (QA) task.}
    \label{tab:qa_t5}
\end{table}

\vspace*{-3mm}

We further validate the performance of \aramus{} by conducting an extensive set of experiments on the ALUE benchmark under few-shot setting. Figure~\ref{fig:fewshot_alue} shows \aramus{} and the best publicly available Arabic PLMs (\jaber{} and \saber{}) performances on 3 representative ALUE tasks (see the full results in Table~\ref{tab:alue_dev_fewshot} of Appendix~\ref{sec:Few-Shot Results}) and the average ALUE score. The 3 selected tasks are: SEC because it shows specific results; OHSD since with FID and OOLD they show similar result patterns, and MDD as a representative of trends observed for tasks MQ2Q, SVREG, and XNLI.


\begin{table}[!ht]
\centering
\small 
\resizebox{\columnwidth}{!}{
    \begin{tabular}{l|ccc}

    \toprule
    & Rouge1 & Rouge2 & RougeL \\
    \midrule
    \multicolumn{4}{c}{\textit{\bf WikiLingua Dev}} \\
    
    \midrule
    \artb & 25.0$\pm$0.2 & 10.0$\pm$0.0 & 22.4$\pm$0.2 \\
    \atb & 26.1$\pm$2.8 & 10.5$\pm$1.6 &  23.2$\pm$2.5 \\

    \midrule
    \aramus & \bf 30.5$\pm$0.1 & \bf 13.2$\pm$0.1 & \bf 26.9$\pm$0.1 \\
    \midrule
    
    \multicolumn{4}{c}{\textit{\bf WikiLingua Test}} \\
    \midrule
    
    \artb & 25.1 & 10.2 & 22.5 \\
    \atb & 27.8 & 11.5 & 24.8 \\
    \midrule
    \aramus & \bf 30.9 & \bf 13.5 & \bf 27.1 \\
    \midrule
    \multicolumn{4}{c}{\textit{\bf EASC Test}} \\
    \midrule
    \artb & 10.7 & 2.7 & 9.3 \\
    \atb & 12.6 & 3.5 & 11.3 \\

    \midrule
    \aramus &  \bf 16.1 &  \bf 6.7 & \bf  13.3 \\
    \bottomrule
    \end{tabular}
    }
    \caption{T5-style models' performances on the Text Summarization task.}
    \label{tab:ts_t5}
\end{table}
\vspace{-3mm}

First, we notice that exceptionally on SEC, \aramus{} performs on par with \jaber{} and underperforms \saber{} on many data points. We think that this is because the text-to-text approach is not effective for multi-label classification tasks under a few-shot setting. Second, we observe that \aramus{} has a marginal gain compared to the best baseline (\saber{}) on some tasks like OHSD, e.g. 0.2\%, 1.0\% and 6.0\% on 8, 128, and 256 examples respectively. As for the remaining 4 tasks (represented by MDD), we observe that \aramus{} significantly outperforms both baselines by a large margin. Overall, \aramus{} shows a consistent performance gain between 4\% to 6\% when averaging the results on the 8 ALUE tasks compared to \saber{}.

\begin{table}[!ht]
\centering
    \begin{tabular}{l|cc}
    \toprule
    \bf Model & \bf Dev & \bf Test \\
    \midrule
    \artb &  6.7$\pm$0.1 &13.5 \\
    \atb & 8.1$\pm$0.1 &  17.0 \\
    \midrule
    \aramus & \bf 8.6$\pm$0.1 & \bf 17.4 \\
    \bottomrule
    \end{tabular}
    \caption{Question Generation dev and test sets BLEU score of T5-style models.}
    \label{tab:qg_emd_t5}
\end{table}
\vspace{-3mm}

Finally, we assess the text generation abilities of \aramus{} by experimenting on 3 generative tasks in Table~\ref{tab:qa_t5},~\ref{tab:ts_t5} and~\ref{tab:qg_emd_t5}. Overall, the observations are consistent with the results obtained on ALUE, \aramus{} reports the highest scores on all tasks and across all metrics. More precisely, \aramus{} significantly outperforms \atb{}, the state-of-the-art Arabic T5-base model, by 
7.5\% and 5.1\% on QA F1 score dev and test sets respectively. Similarly, \aramus{} has a 
gain of 4.4\%, 4.1\%, and 3.5\% on TS dev, test, and EASC test rouge1 score respectively. However, gains are not always significant on generative tasks, as we observe a smaller margin of improvement of 0.5\% and 0.4\% and against the best baseline on QG dev and test sets respectively.

\section{Conclusion}
\label{sec:Conclusion}
\vspace*{-3mm}

In this paper, we introduced \aramus{} which is not only the largest Arabic PLM in terms of pre-training data and model size, but also the first multi-billion Arabic PLM to be extensively evaluated on a wide range of NLP tasks. Since our work gives clues on the benefits and limitations of scaling up data and model sizes, we hope that it will pave the way for the Arabic NLP community to focus on problems that are beyond the reach of PLM scaling.

\section*{Limitations}
\vspace*{-2mm}

While our model shows state-of-the-art results on many discriminative and generative tasks, we can think of the following main caveats of our work. First, the number of generative tasks that we evaluate on is relatively small especially when consider that \aramus{} is text-to-text encoder-decoder model. This is mainly because of the rarity of Arabic generative datasets that are at the same time well-established and open-source. Second, it would be important to study how end-tasks performances is impacted when ablating the model size (e.g. 1-6 billion parameters models), pretraining data quantity or/and quality.

\normalem 
\bibliography{aramus}
\bibliographystyle{acl_natbib}

\clearpage
\newpage
\appendix

\section{Pretraining}
\vspace*{-2mm}

\subsection{Data Collection}
\label{sec:Pretraining Data Collection}
\vspace*{-1mm}
Our pre-training corpus is mainly sourced from the publicly available web scrapes of the Common Crawl (CC) project. We downloaded 90 shards of CC monthly data ranging from May 2013 (the earliest available) up to July 2022. Also, we use an \textit{in-house collection} of 47GB of Arabic dialect textual data (DIALECT) in order to enhance the awareness of our model to Arabic dialects~\cite{abdul2021arbert}. In addition, we include high-quality news corpora such as NEWS~\cite{zeroual2019osian} and El-KHAIR~\cite{el20161} which are commonly used in previous Arabic PLM works~\cite{safaya2020kuisail,antoun2020arabert,nagoudi2022_arat5,ghaddar2022revisiting}. Finally, we use 28GB of \textit{in-house} Arabic data curated from different text genres like literature, books, and Wikipedia.

\begin{table}[!ht]
\centering
\small

\begin{tabular}{l|r|r|c}
\toprule
\bf Source & \bf Original & \bf Clean   & \bf Filtering \%\\
\midrule
 CC  & 8.7TB & 439GB & 95\%  \\
 DIALECT  & - & 47GB &  -  \\ 
 NEWS  & 21GB & 14GB &  34\% \\
 EL-KHEIR  & 16GB & 13GB &  19\% \\
 Others  & 28GB & 16GB &  45\%\\
 \midrule
 Total  & 8.8TB & 529GB &  94\% \\
\bottomrule
\end{tabular}

\caption{Size of the pre-training corpora before (Original) and after (Clean) applying data filtering and deduplication heuristics.}
\label{tab:pretrain_data}
\end{table}

\vspace*{-3mm}
As it has been shown to be crucial for English~\cite{raffel2019exploring}, multilingual~\cite{xue2021mt5}, and Arabic~\cite{ghaddar2022revisiting} PLM end-tasks performance, we aggressively filter and deduplicate the collected data using the heuristics described in ~\cite{ghaddar2022revisiting}. Table \ref{tab:pretrain_data} shows data sizes before and after applying the heuristics. While we discard 95\% of CC data, it is still considered, along with DIALECT, to form more than 90\% of our 529GB final pre-training corpus. 

\subsection{Implementation Details}
\label{sec:Pretraining Implementation Details}
\vspace*{-1mm}

We use the SentencePiece~\cite{kudo2018sentencepiece} tokenizer in order to process text into sub-tokens. We train the tokenizer from scratch on our pre-training corpus by setting  the vocabulary size to 64k, a value which is used commonly by previous Arabic PLMs  ~\cite{antoun2020arabert,ghaddar2022revisiting,nagoudi2022jasmine}.

Following~\cite{raffel2019exploring}, we pre-train \aramus{} on the \textit{Replace corrupted spans} tasks with a random token probability of~15\%. The pre-training code is based on the PyTorch~\cite{paszke2019pytorch} version of the Megatron-LM library~\cite{shoeybi2019megatron}. \aramus{} is pre-trained on 16 sever, each occupied with 8 NVIDIA A100 GPUs with 80GB memory. Model and data parallel sizes are set to 4 and 32 respectively. The total batch size is 4096, which is based on the max batch size which can fit on a single GPU (32).  To speed up the pre-training, 
we use mixed-precision training~\citep{fp16}, except when calculating attention softmax and when reducing gradients. We use the Adafactor optimizer~\cite{adafactor2018} with an initial learning rate of $0.005$, 10k warm-up steps with the inverse square-root scheduler.

\section{Finetuning}
\label{sec:Finetuning}
\vspace*{-2mm}

\subsection{Datasets and Evaluation}
\label{sec:Finetuning:Datasets and Evaluation}
\vspace*{-1mm}

ALUE~\cite{seelawi2021alue} is a well-established benchmark that consists of a collection of eight Arabic NLU tasks. Although its datasets are relatively small compared to the one of the English GLUE~\cite{wang2018glue} benchmark, but it is supported by a public leaderboard with hidden test sets which ensures a fair comparison between models. Following~\cite{seelawi2021alue}, we report Pearson correlation on SVREG, Jaccard on  SEC, and accuracy on XNLI, and use the F1 score otherwise. We also report the unweighted average sum over the 8 tasks.

As for generative tasks, we follow ~\cite{ghaddar2022revisiting} by considering 3 tasks for evaluation, as their datasets are fully open source. We use Wikilingua~\cite{ladhak-etal-2020-wikilingua} and EASC~\cite{ElHaj2010UsingMT} for TS, and the set of datasets used in ~\cite{abdul2021arbert,nagoudi2022_arat5} for QA and QG. We follow~\cite{ghaddar2022revisiting} for splitting the data into train/dev/test, and report Rouge scores~\cite{lin2004rouge} on TS, BLEU~\cite{papineni2002bleu} on QG, and Exact Match (EM) and F1 score on QA. Therefore, \aramus{} results can be directly comparable with the baselines reported by~\cite{ghaddar2022revisiting}. 

\subsection{Baseline}
\label{sec:Baseline}
\vspace*{-1mm}

We compared \aramus{} with the state-of-the-art Arabic PLMs that have been evaluated on publicly available datasets, these include:
\vspace{-2mm}
\begin{itemize}
	
    \item \textbf{\arbert{} and \marbert{}} are respectively MSA and Arabic Dialect \bert{-base}~\cite{devlin2018bert} models provided by ~\cite{abdul2021arbert}.
	\vspace*{-3mm}
    \item \textbf{\jaber{} and \saber{}} are  respectively \bert{-base} and \bert{-large} models provided by \cite{ghaddar2022revisiting}. 
    \vspace*{-3mm}
    \item \textbf{ALM-1.0}~\footnote{\url{https://github.com/FlagAI-Open/FlagAI/tree/master/examples/ALM}} is a recently published Arabic \bert{-large} model. 
    \vspace*{-3mm}
    \item \textbf{\artb{} and \atb{}} are Arabic T5-base~\cite{raffel2019exploring} models provided by ~\cite{nagoudi2022_arat5} and \cite{ghaddar2022revisiting} respectively.

	
\end{itemize}
\vspace{-2mm}
It is worth mentioning that it was not possible to compare \aramus{} with its counterpart multi-billion Arabic GPT models because:

\subsubsection{NOOR}
\label{Noor}
\vspace*{-1mm}
NOOR~\cite{lakim2022holistic} is the largest existing Arabic PLM with 10B parameters. In their work, the authors didn't make their model publicly available neither reported their results on public datasets.  

\subsubsection{AraGPT-Mega}
\label{AraGPT}
\vspace*{-1mm}
AraGPT-Mega~\cite{antoun2021aragpt2} has  1.5B parameters and is publicly available for download. However, we tried to run \textit{in-house} experiments with this model but it didn't perform well on many tasks. Most likely because it was only pre-trained on 27GB of Arabic text, which is considered small compared to the model size. Therefore, we preferred not to report weak results for this model.

\subsubsection{Jasmine}
\label{Jasmine}
\vspace*{-1mm}
Jasmine~\cite{nagoudi2022jasmine} is an \textit{in-progress} project that aims to develop and evaluate a set of Arabic GPT models up to 13B parameters. This in-progress work was released at the time of writing our paper.  
The authors mentioned that the 13B model is still at early pre-training stage, while the 6.7B version is only pre-trained for 143k steps. Therefore, their \textit{fully pre-trained} Jasmine has 2.7B parameters only. This model is evaluated, in a few shot setting only, on a set of discriminative and generative tasks on the ARLUE~\cite{abdul2021arbert} and ARGEN~\cite{nagoudi2022_arat5} benchmarks respectively. However, many of the datasets in ARLUE and ARGEN have not been publicized yet~\cite{elmadany2022orca,ghaddar2022revisiting}. In addition, the authors didn't open source their model weights nor shared their code to replicate their dataset splits.

\subsection{Implementation Details}
\vspace*{-1mm}

We used early stopping based on the performance of the dev sets during our extensive hyper-parameter search. We search the learning rate from the set of \{5e-5, 1e-4, 2e-4, 1e-3\}, batch size from \{8, 16, 32, 64\}, the learning rate scheduler from \{constant, cosine\}, and the dropout rate from  \{0.1, 0.15, 0.2, 0.3\}, and fixed the epoch number to a maximum of 120 for all the experiments. Each fine-tuning experiment uses 4 NVIDIA A100 GPUs, with the model parallel size set to 4. 

After finding the best hyper-parameters, we ran all the experiments 5 times and reported the average score on the dev sets~\footnote{We use the MQ2Q dev set curated by~\cite{ghaddar2022revisiting} to make our results compatible with their baselines.}, in order to validate the credibility of our results. For each ALUE task, we selected the best-performing model among the 5 runs and used it for the ALUE leaderboard test submission, and we computed the scores on generative tasks datasets. 

We simulate a few-shot setting on the ALUE tasks by randomly sampling a subset of \{8, 16, 32, 64, 128, 256\} examples of the training data. When the number of classes is more than the number of samples (e.g. MDD and SEC with 8 examples) we randomly add one example for each missing class in order to ensure that each class has a represented data point. All models are identically fine-tuned, and we report the average and standard deviation of 5 randomly selected folds.

\section{Few-Shot Results}
\label{sec:Few-Shot Results}
\vspace{-5cm}

\begin{table*}[t]
\small
\centering
\resizebox{\textwidth}{!}{
\begin{tabular}{@{}lccccccccc@{}}
\toprule

\textbf{Model} & \textbf{MQ2Q*} & \textbf{MDD} & \textbf{SVREG} & \textbf{SEC}& \textbf{FID}& \textbf{OOLD}& \textbf{XNLI}& \textbf{OHSD} & \textbf{Avg.}\\
\midrule
\small\textit{8 Examples} & & & & & & & & &\\
\midrule
\jaber & 50.0$\pm$15.8 & \underline{8.9$\pm$1.8} & 18.8$\pm$17.5 & \underline{21.7$\pm$0.2} & 56.7$\pm$13.5 & 56.5$\pm$7.9 & 35.7$\pm$2.3 & 54.9$\pm$5.9 & 37.9$\pm$8.1 \\
\saber & 53.5$\pm$6.9 & \underline{8.8$\pm$1.4} &  34.2$\pm$09.6 & \underline{21.1$\pm$0.8} & 63.0$\pm$11.2 & 65.3$\pm$12.6 & 35.5$\pm$1.9 & 58.1$\pm$7.3 & 42.4$\pm$6.5 \\
\aramus & \bf 60.2$\pm$3.7 & \underline{\bf 16.7$\pm$1.8} & \bf 54.5$\pm$8.7 & \underline{\bf 23.2$\pm$3.5} & \bf 69.0$\pm$2.8 & \bf 69.5$\pm$1.6 & \bf 35.8$\pm$1.1 & \bf 58.3$\pm$7.7 & \bf 48.4$\pm$3.9 \\

\midrule
\small\textit{16 Examples} & & & & & & & & &\\
\midrule
\jaber & 56.2$\pm$14.5 & \underline{7.9$\pm$1.1} & 45.2$\pm$16.1 & \underline{24.3$\pm$3.0} & 69.9$\pm$5.6 & 68.0$\pm$12.5 & 37.0$\pm$3.4 & 53.0$\pm$5.7 & 45.2$\pm$7.7 \\
\saber & 54.6$\pm$8.2 & \underline{9.0$\pm$2.1} & 47.7$\pm$16.7 & \underline{21.6$\pm$1.9} & 73.0$\pm$2.8 & 80.3$\pm$7.9 & 35.8$\pm$2.3 & 57.7$\pm$8.0 & 47.5$\pm$6.2 \\
\aramus & \bf 61.4$\pm$4.7 & \underline{\bf 20.4$\pm$1.9} & \bf 66.6$\pm$5.6 & \underline{\bf 25.5$\pm$4.8} & \bf 74.3$\pm$1.2 & \bf 82.3$\pm$1.7 & \bf 39.1$\pm$4.9 & \bf 59.1$\pm$7.5 & \bf 53.6$\pm$4.0 \\
\midrule
\small\textit{32 Examples} & & & & & & & & &\\
\midrule

\jaber & 66.9$\pm$3.3 & \underline{8.0$\pm$1.8} & 63.7$\pm$11.7 & 27.0$\pm$3.3 & 72.1$\pm$3.9 & 71.7$\pm$5.9 & 38.7$\pm$2.9 & 57.7$\pm$7.7 & 50.7$\pm$5.1 \\
\saber & 63.3$\pm$6.6 & \underline{9.8$\pm$2.3} & 72.3$\pm$9.3 & 28.7$\pm$4.1 & 74.5$\pm$1.4 & 81.2$\pm$9.3 & 37.4$\pm$1.4 & 54.6$\pm$7.2 & 52.7$\pm$5.2 \\
\aramus & \bf 69.2$\pm$4.3 & \underline{\bf 21.3$\pm$1.1} & \bf 74.5$\pm$3.6 & 28.0$\pm$5.0 & \bf 74.8$\pm$2.6 & \bf 85.5$\pm$2.1 & \bf 45.3$\pm$3.7 & \bf 59.3$\pm$6.9 & \bf 57.2$\pm$3.7 \\

\midrule
\small\textit{64 Examples} & & & & & & & & &\\
\midrule

\jaber & 68.6$\pm$3.5 & 11.0$\pm$1.9 & 72.6$\pm$7.8 & 31.5$\pm$1.5 & 73.7$\pm$0.8 & 77.0$\pm$2.7 & 42.4$\pm$2.2 & 58.8$\pm$8.4 & 54.4$\pm$3.6 \\
\saber & 67.8$\pm$2.8 & 12.8$\pm$1.9 & 79.6$\pm$3.3 & 34.8$\pm$1.7 & 77.2$\pm$1.4 & 87.0$\pm$2.1 & 39.6$\pm$4.2 & 61.4$\pm$7.4 & 57.5$\pm$3.1 \\
\aramus & \bf 74.8$\pm$1.8 & \bf 22.3$\pm$1.0 & \bf 81.8$\pm$3.7 & 31.5$\pm$1.8 & \bf 77.7$\pm$0.7 & \bf 89.6$\pm$1.5 & \bf 55.5$\pm$3.6 & \bf 64.0$\pm$8.7 & \bf 62.2$\pm$2.8 \\

\midrule
\small\textit{128 Examples} & & & & & & & & &\\
\midrule
\jaber & 70.0$\pm$1.5 & 16.9$\pm$0.6 & 80.5$\pm$1.3 & 35.3$\pm$1.8 & 76.4$\pm$1.1 & 82.4$\pm$2.8 & 44.6$\pm$1.0 & 64.2$\pm$4.0 & 58.8$\pm$1.8 \\
\saber & 72.1$\pm$0.9 & 18.9$\pm$2.0 & 83.6$\pm$2.0 & 39.5$\pm$2.8 & 78.3$\pm$1.3 & 88.7$\pm$1.4 & 44.8$\pm$4.0 & 66.8$\pm$4.0 & 61.6$\pm$2.3 \\
\aramus & \bf 77.5$\pm$1.1 & \bf 25.7$\pm$1.7 & \bf 84.1$\pm$0.9 & 37.0$\pm$1.4 & \bf 78.6$\pm$0.5 & \bf 90.4$\pm$0.9 & \bf 63.6$\pm$1.5 & \bf 67.8$\pm$4.1 & \bf 65.6$\pm$1.5 \\
\midrule
\small\textit{256 Examples} & & & & & & & & &\\
\midrule

\jaber & 72.7$\pm$1.0 & 22.4$\pm$0.6 & 83.7$\pm$0.7 & 39.3$\pm$0.8 & 79.0$\pm$1.1 & 84.9$\pm$1.0 & 53.1$\pm$2.2 & 62.5$\pm$6.2 & 62.2$\pm$1.7 \\
\saber & 72.8$\pm$1.7 & 25.5$\pm$1.9 & 85.0$\pm$1.3 & 42.2$\pm$0.5 & 79.8$\pm$1.2 & 89.6$\pm$0.7 & 48.0$\pm$13.5 & 70.6$\pm$1.3 & 64.2$\pm$2.8 \\
\aramus & \bf 78.1$\pm$1.2 & \bf 30.2$\pm$0.8 & \bf 86.3$\pm$1.3 & 41.1$\pm$0.7 & \bf 80.8$\pm$1.7 & \bf 92.3$\pm$0.9 & \bf 72.6$\pm$0.7 & \bf 71.2$\pm$3.4 & \bf 69.1$\pm$1.3 \\
\bottomrule

\end{tabular}
}
\caption{Dev ALUE performances across training set sizes. Underline figures indicates extra samples where added to ensure that each class is represented  at least by one data point.}
\label{tab:alue_dev_fewshot}
\end{table*}

\end{document}